\documentclass[11pt]{article}
\usepackage{eamt22}
\usepackage{times}
\usepackage{url}
\usepackage{latexsym}
\usepackage[small,bf]{caption} 
\usepackage{multirow}
\usepackage{adjustbox}
\usepackage{layout}
\usepackage{subfigure}
\usepackage{amsmath}
\usepackage{enumitem}
\usepackage[table]{xcolor}

\newcommand{\conffilename}{eamt22}

\title{A Systematic Analysis of Vocabulary and BPE Settings for Optimal Fine-tuning of NMT: A Case Study of In-domain Translation}

\author{Javad Pourmostafa Roshan Sharami, Dimitar Shterionov, and Pieter Spronck\\
  Department of Cognitive Science and Artificial Intelligence, Tilburg University, Tilburg, The Netherlands\\
  {\tt \{j.pourmostafa,d.shterionov,p.spronck\}@tilburguniversity.edu}}  

\date{}

\begin{document}
\maketitle
\begin{abstract}
The effectiveness of Neural Machine Translation~(NMT) models largely depends on the vocabulary used at training; small vocabularies can lead to out-of-vocabulary problems -- large ones, to memory issues. Subword (SW) tokenization has been successfully employed to mitigate these issues. The choice of vocabulary and SW tokenization has a significant impact on both training and fine-tuning an NMT model. Fine-tuning is a common practice in optimizing an MT model with respect to new data. However, new data potentially introduces new words (or tokens), which, if not taken into consideration, may lead to suboptimal performance. In addition, the distribution of tokens in the new data can differ from the distribution of the original data. As such, the original SW tokenization model could be less suitable for the new data. 

Through a systematic empirical evaluation, in this work we compare different strategies for SW tokenization and vocabulary generation with the ultimate goal to uncover an optimal setting for fine-tuning a domain-specific model. Furthermore, we developed several (in-domain) models, the best of which achieves $6$ BLEU points improvement over the baseline.

\end{abstract}

\section{Introduction}
\label{sec:introduction}



Fine-tuning is a common practice in optimizing an MT model with respect to new data. It can be either in the context of domain adaptation where an existing model is tuned for a certain domain (different from what the model was originally trained for)~\cite{dakw:17fine,70c446486deb46eba84b4e2d06c5f963,mahdieh2020rapid}, or simply to improve the performance of the model~\cite{wang-etal-2017-sentence}. Regardless of the fine-tuning objective, the newly introduced data brings in new information. For example, it could be the case that the new data contains new words that have not been seen in previous cycles of training the model; the word segmentation is suboptimal for the new data~\cite{lim2018exploring,essay80128,sato-etal-2020-vocabulary}. If not properly addressed, this new information may not have the desired effect on the MT system. For example, if new words are not included in the vocabulary a mismatch in vocabulary causing an out-of-vocabulary (OOV) issues will occur.

In this paper, we present an empirical evaluation of several models trained on different combinations of 
settings for generating the subwords and vocabularies, aiming to identify a best-case setup for fine-tuning an MT engine. 
That is, we aim to investigate which fine-tuning conditions (or settings) of a domain-specific model lead to the best performance. 
Our research focuses on fine-tuning for improving the translation quality of an engine rather than on domain adaptation.
With our research we aim to answer the following main research question:

\begin{enumerate}[leftmargin=0.9cm]
    \item[RQ1] Given a machine translation model and a fine-tuning data set, what is the optimal combination of subword generation approach and vocabulary?
\end{enumerate}

In addition, we investigate two secondary research questions:
\begin{enumerate}[leftmargin=0.9cm]
    \item[RQ2] How does fine-tuning, i.e. training an MT system on one data set, followed by training on another, compare to training an MT system with all data at once?
    \item[RQ3] What is the time reduction, if any, when fine-tuning, compared to training an MT system with all data at once?
\end{enumerate}

To answer these questions we exploit two data sets ($\sim$0.9M and $\sim$1.1M parallel sentences) to train and one data set ($\sim$179k parallel sentences) to fine-tune several MT systems. This we do so that we can investigate how much we can improve a model trained on sufficient amount of data (approximately $1M$ sentence pairs) and then fine-tuned with extra data, which on its own would not be enough to train a model.     

Each fine-tuned alternative, is trained on a different set of options of how the subwords and the vocabulary are created. 
For the fine-tuning process, we proposed a method to find the best fine-tuning setup based on available data. In our case study, for example, we have access to the data of both models (initial and fine-tuned), however, as already discussed in~\cite{freitag2016fast,dakw:17fine}, initial models are mostly deployed in an application; thus data might not be available at the production time. As such, it is paramount to have a guideline based on the available data THAT determines how to best generate sub-words and vocabularies. In this work we use Byte-Pair Encoding~\cite{sennrich-etal-2016-controlling} for subword units.

We also trained models with all available data at once to assess whether fine-tuning has any benefit. We also evaluated MT systems trained on the small in-domain data set first and then fine-tuned on the larger sets. That is in order to test the hypothesis that a system trained on a small, focused data set and then fine-tuned on a larger set is worse than the other way round.\footnote{This hypothesis has been proven in other domains, e.g. in robots control systems~\cite{Spronck2008_DECA}.} 

It is noteworthy that the point of this research, however, is to investigate what is the best fine-tuning setup; and not to find the best model. That is, if we start from data set A (regardless of whether it is in-domain, out-of-domain, selected, or other) and then we fine tune on data set B, what should we take under consideration with respect to BPE and vocabulary.

This paper is organized as follows. We first cover the data we used in our experiments. In Section~\ref{sec:methodology}, we present our decision points. Our empirical experiments including details on the data, subwords, vocabulary, systems specifications, baselines and results are shown in Section~\ref{sec:experiments}. Section~\ref{sec:discussion} presents our analysis with respect to the RQs. We cover the related work and conclude our work in Section~\ref{sec:related_work} and \ref{sec:conclusion}, respectively.

\section{Data}
\label{sec:data}

In our research we used two data sets: (i) two corpora of \textit{selected in-domain data} in which sentence pairs have been selected from an out-of-domain corpus, i.e., $\sim$0.9M and $\sim$1.1M selected from the $\sim$31M sentences of collected WMT corpora\footnote{\url{http://statmt.org/wmt15/translation-task.html}} according to the data selection method presented in~\cite{Pourmostafa_Roshan_Sharami_Sterionov_Spronck_2022}; and (ii) a small ($\sim$179K parallel sentences) \textit{original in-domain data} set. 

\paragraph{Selected In-domain Data set} 
\label{CLIN_Paper}
The selected in-domain data used for training the initial models (i.e., before doing fine-tuning), was introduced in~\cite{Pourmostafa_Roshan_Sharami_Sterionov_Spronck_2022}. We used, in particular, $Top5$ and $Top6$ because they led to the best translation performance in their work. The selected in-domain data is the result of ranking out-of-domain sentences according to their similarity with an in-domain data set. The language pair is English-French. Furthermore, as indicated by their work, the volume of the data is sufficient to train MT systems with high translation quality.

\paragraph{Original In-domain Data set} 
\label{original_in_domain}
The original in-domain data we experimented with is the International Workshop on Spoken Language Translation (IWSLT) 2014 corpus~\cite{Cettolo2015ReportOT}. It is a collection of TED talks. To evaluate our models during training and find the models' performance we used one development set (dev2010) and two test sets (test2010 and test2011), respectively. IWSLT 2014 and WMT are commonly used in the context of Domain Adaptation (DA) as an in-domain data set~\cite{axelrod-etal-2011-domain,Luong2015StanfordNM,Chen2016BilingualMF,wang-etal-2017-sentence,Pourmostafa_Roshan_Sharami_Sterionov_Spronck_2022}, which facilitates for better replicability. 

Table~\ref{tab:corpora} shows statistics of the data we used in our experiments. 

\begin{table}[ht]
\renewcommand{\arraystretch}{1.5}
\centering
{\small
\begin{tabular}{|c|c|c|}
\hline
\hline
\textbf{Type of data}               & \textbf{Name}                  & \textbf{Sentences}   \\ \hline
\multirow{2}{*}{\begin{tabular}[c]{@{}c@{}}Selected in-domain\\ \end{tabular}} & Top5                           & 895k                 \\ \cline{2-3} 
                                    & Top6                           & $\sim$1M             \\ \hline 
\multirow{4}{*}{\begin{tabular}[c]{@{}c@{}}Original in-domain\\ (IWSLT 2014)\end{tabular}} & TED training       & 179K                \\ \cline{2-3} 
              & TED dev2010                              & 887                 \\ \cline{2-3} 
             & TED test2010                             & 1664                \\ \cline{2-3} 
             & TED test2011                             & 818                 \\ \hline \hline
\end{tabular}}
\caption{Summary of in-domain data sets~(in-domain and out-of-domain), plus out-of-domain data sets.}\label{tab:corpora}
\end{table}

\section{Decision points}
\label{sec:methodology}
As noted in Section~\ref{sec:introduction}, with this work we aim to identify the most effective way of fine-tuning an MT system with respect to \emph{subwords} and \emph{vocabulary}.
Consider a model $M$ trained on a data set $D$ which is representative for a certain domain $d$ and a fine-tuning data set $E$. The following decision points need to be made:

\paragraph{Available data:} Choose which fine-tuning data set or a combination of data sets from $E$ should be used. As shown in previous work, using all available data is not always beneficial as it does not always contribute to the overall performance (especially when it comes to specific domains) while introducing computational overhead~\cite{70c446486deb46eba84b4e2d06c5f963,soto-etal-2020-selecting,Pourmostafa_Roshan_Sharami_Sterionov_Spronck_2022}.

\paragraph{Subwords:} Choose a model to construct subword units. Use either (i) the BPE model learned on the set $D$, and thus used in the training of model $M$ ($D_{BPE}$), (ii) learn a new BPE model on the selected fine-tuning data set ($E_{BPE}$) or (iii) learn a new BPE on the concatenation of $D$ and $E$ ($(D+E)_{BPE}$). This is mainly because there might be new and unique words in the fine-tuning data that did not appear in the set $D$, for which the original BPE model would be suboptimal. That can be the case if we tune an existing model toward a different domain other than $d$. However, comparing (ii) and (iii), training data from the original model may not be available and as such only (ii) could be a viable option.
    
\paragraph{Vocabulary:} Choose the vocabulary, that is, either (i) use the vocabulary of the original model $M$ ($|D|$), (ii) extend it with tokens from the fine-tuning set $E$ ($|D+E|$) or (iii) create a new vocabulary from $E$ ($|E|$). This is important because a relevant vocabulary set is the pillar of the MT performance. Thus, finding such a set mitigates the impact of OOV and rare words.


These two factors -- subwords and vocabulary -- need to be jointly considered as each of them has a significant impact on the MT performance. 
To this end, we face an optimization problem along two dimensions. As the different options at each dimension are independent of the rest,~\footnote{It is noteworthy that we are aware that vocabulary sets in every combination are dependent on BPE models, and here we only defined ``combinations'' as independent compared to their other peers.} the solutions can be enumerated as combinations over these options. 
This gives us 9 combinations: apply the BPE model of the original MT system $D_{BPE}$ on the data for fine-tuning $E$ and train three systems with the different vocabulary options $|D|$, $|D+E|$ and $|E|$; train a new BPE model on the fine-tuning data, $E_{BPE}$ and apply it on the data, generating three different vocabularies ($|D|$, $|D+E|$ and $|E|$). 


Following these decision points, given a fine-tuning data set we can consider three BPE models. With these models we (i) \textit{create the vocabulary sources}; and (ii) \textit{create the training sets for fine-tuning}. Typically these two processes are tied to each other, i.e. once the BPE model is learned and applied on the training data, the vocabulary is the set of subword units that appear in the (processed) data. However, this is not a hard constraint. That is, we can use a vocabulary that is derived from data processed with a different BPE model than the one of the training or fine-tuning data. For instance, data set $E$ can be processed with BPE $E_{BPE}$ but the vocabulary used for training can still be based on $D$ derived from applying $D_{BPE}$. 



\section{Experiments}
\label{sec:experiments}
To find the best possible setting for fine-tuning an in-domain model, we followed the decision points in Section~\ref{sec:methodology}, conducted experiments with the English-French data presented in Section~\ref{sec:data} and compared our fine-tuned models with each other and to different baselines using BLEU~\cite{Papineni2002}, TER~\cite{snover-etal-2006-study} and chrF~\cite{popovic-2015-chrf}. 

\begin{figure}[ht]
	\centering 
	\includegraphics[width=\linewidth]{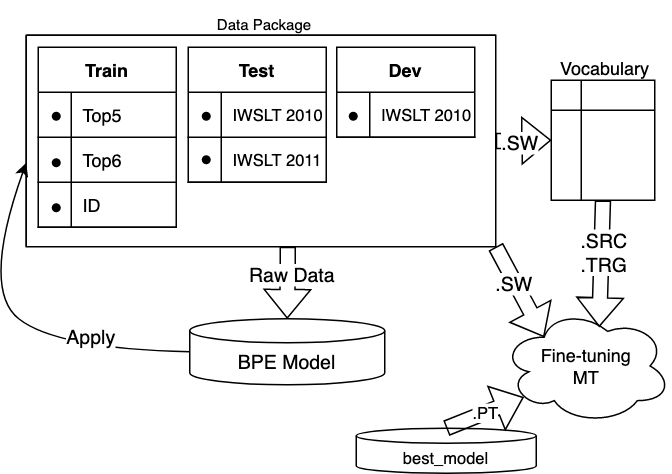} 
	\caption{An overview of fine-tuning an MT. The data package shows the data sets we used for experiments. These can be used for training the initial models or fine-tuning the trained models.} 
	\label{fig:overview}
\end{figure}

Figure~\ref{fig:overview} illustrates our experimental approach. 

According to the given data and our task, i.e., fine-tuning a model trained on $Top5$ or $Top6$ ($D = Top5$ or $D = Top6$) with original in-domain data ($E = $in-domain data) we have the 54 options, 27 for each trained model. There are 3 options for the data source of the vocabulary ($Top5$ or $Top6$, $ID$ or the combination thereof) and 3 options for building the BPE which then impact the segmentation of the data used to build the vocabulary but also the segmentation of the fine-tuning data.\footnote{As noted at the end of Section~\ref{sec:methodology} we are not strictly constrained against mixing different word segmentations for the training data and for the vocabulary in a fine-tuning test case.} That is, there are 3 options to build a BPE model to be used for segmenting the data from which the vocabulary will be created and 3 options to build a BPE model to be used to segment the fine-tuning data. 

It is noteworthy that the number of merge operations for BPE is 50K, and a separate BPE model was created for each source and target.

\begin{figure*}[ht]
	\centering 
	\includegraphics[width=0.95\textwidth]{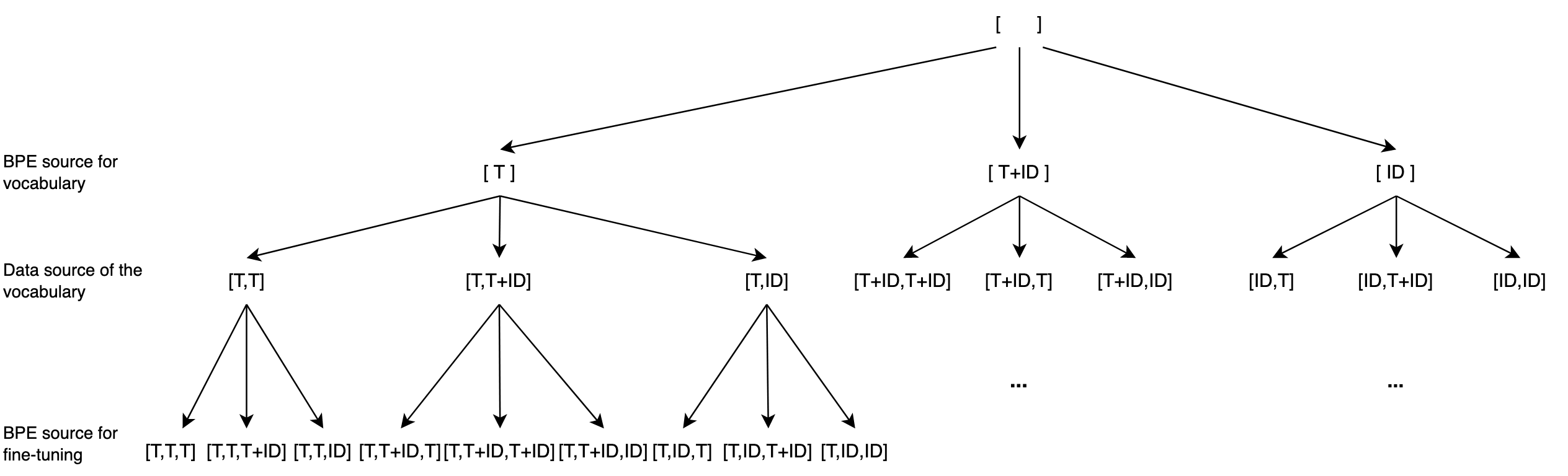} 
	\caption{Tree's leaves show different combinations that we experimented with. For sake of simplicity in our report, we assume $Top5$ and $Top6$ as one single training data. That is, we experimented with both $Top5$ and $Top6$, however, we did not expand the tree for each of them. ``$T$'' also abbreviated from ``$Top$''.} 
	\label{fig:tree}
\end{figure*}

Figure~\ref{fig:tree} gives an overview of the different options.

Based on these 54 options we define three types of experiments. First, experiments in which the BPE model is built on either the original data $D = Top5/6$ or on the fine-tuning data $E = ID$ (but not on their combination) and the vocabulary is generated from the same data. With this type of experiments we investigate (hypothetical) scenarios where either there is no access to the original data set ($D$) nor vocabulary ($|D|$), and as such only the fine-tuning dataset ($E$) can be used, or these are available and we can exploit them directly without spending time or resources on processing the fine-tuning data to extract $|E|$. Second, experiments in which the vocabulary is built on both $D$ and $E$ ($Top5/6 + ID$). This would be considered a very favourable scenario, where both $D$ and $E$ are available and can be exploited jointly. Under such assumption, we also build baseline models on $D + E$ (see Section~\ref{sec:baselines}). Third, experiments in which the BPE used to segment the data on which the vocabulary is built is different from the BPE used to segment the fine-tuning data. These experiments would cover (hypothetical) scenarios in which the vocabulary is given, but it does not correspond to the exact way the subwords of the fine-tuning data have been generated. With this, third, set of experiments, we want to see whether it is possible to reach sufficient quality under limiting conditions. 

To reduce the amount of computational time and resources we implemented 11 of the 27 sets of experiments. For the second type of experiments we excluded those in which different BPE models are used for the vocabulary and for the fine-tuning data. That is because given that both $D$ and $E$ are mixing BPE models would be unnecessary and impractical. Following the same reasoning, experiments from type 1 and 3 where two BPE models -- one learned from $D + E$ and another learned either from $D$ or from $E$ -- were excluded. These leaves us with the 11 experiments enumerated in Table~\ref{tbl:result}.

\subsection{NMT System Description}
We used the OpenNMT-py\footnote{\url{https://opennmt.net/OpenNMT-py/}} framework~\cite{klein-etal-2017-opennmt} for training as well as fine-tuning our NMT models. We fine-tuned transformer models~\cite{vaswani2017attention} for a maximum of 200K steps; intermediate models were saved and validated every $1000$ steps until reached convergence. We set an early stopping condition such that the fine-tuning process was stopped after $10$ validations steps with no improvement. Since the models we fine-tuned were proposed by \cite{Pourmostafa_Roshan_Sharami_Sterionov_Spronck_2022}, we kept the NMT setup consistent and used the same hyperparameters. To run all NMT systems effectively and aligned with our research questions, we also set other hyperparameters as suggested by the OpenNMT-py community to simulate Google's default setup~\cite{vaswani2017attention}.

For fine-tuning, we distributed the training over three NVIDIA Tesla V100 GPUs. We encoded all data as a sequence of subwords units using the Byte Pair Encoding (BPE) algorithm~\cite{sennrich-etal-2016-controlling}. The number of BPE merge operations is 50K, and a separate BPE model was created for each source and target.

\subsection{Baseline Models} \label{sec:baselines}
We compared our fine-tuned models not only to each other but also to three different categories of baselines: 

(1)~We only trained NMT systems on the original or in-domain data (with no further training/fine-tuning) to set up the following baselines. B1 -– an NMT model trained on the original in-domain data; B2 and B3 -– NMT models trained on the selected in-domain data~Top5 and Top6 respectively. As stated before, we did not fine-tune any models to establish B1, B2, and B3, however, we use them as baselines to measure fine-tuning improvement over the IWSLT data set. 

(2)~We considered several state-of-the-art fine-tuned models. These are compared with our models to show the impact of our fine-tuning process. These are B4~\cite{Luong2015StanfordNM}, B5~\cite{axelrod-etal-2011-domain}, B6~\cite{Chen2016BilingualMF}, B7~\cite{wang-etal-2017-sentence}. It is worth mentioning that we defined these baselines to evaluate the impact of the fine-tuning procedure itself before comparing the proposed combinations with each other. 

(3)~We mixed original ($Top5$ or $Top6$) and in-domain ($ID$) data sets and then trained a model. That is, we did not fine-tune any models in this category. This helps to show and measure the difference between merge and fine-tuning operations. 

The baseline results are shown in Table~\ref{tbl:baselines}.  

\begin{table}[ht]
\renewcommand{\arraystretch}{1.5}
\centering
\begin{adjustbox}{width=\linewidth,center}
\begin{tabular}{|c|ccc|ccc|}
\hline \hline
\multirow{2}{*}{\#} & \multicolumn{3}{c|}{Test set 2010}                            & \multicolumn{3}{c|}{Test set 2011}                            \\ \cline{2-7} 
                   & \multicolumn{1}{c|}{BLEU$\uparrow$} & \multicolumn{1}{c|}{TER$\downarrow$}  & CHRF2$\uparrow$ & \multicolumn{1}{c|}{BLEU$\uparrow$} & \multicolumn{1}{c|}{TER$\downarrow$}  & CHRF2$\uparrow$ \\ \hline
B1 & \multicolumn{1}{c|}{31.9} & \multicolumn{1}{c|}{56.6} & 57.0 & \multicolumn{1}{c|}{38.3} & \multicolumn{1}{c|}{49.7} & 61.0 \\ \hline
B2          & \multicolumn{1}{c|}{30.9} & \multicolumn{1}{c|}{59.1} & 57.0  & \multicolumn{1}{c|}{36.7} & \multicolumn{1}{c|}{51.5} & 62.0  \\ \hline
B3          & \multicolumn{1}{c|}{31.3} & \multicolumn{1}{c|}{58.3} & 58.0  & \multicolumn{1}{c|}{36.5} & \multicolumn{1}{c|}{50.9} & 62.0  \\ \hline \hline
B4          & \multicolumn{1}{c|}{32.2} & \multicolumn{1}{c|}{N/A} & N/A  & \multicolumn{1}{c|}{35.0} & \multicolumn{1}{c|}{N/A} & N/A  \\ \hline
B5          & \multicolumn{1}{c|}{32.2} & \multicolumn{1}{c|}{58.3} & N/A  & \multicolumn{1}{c|}{35.5} & \multicolumn{1}{c|}{N/A} & N/A  \\ \hline
B6          & \multicolumn{1}{c|}{30.3} & \multicolumn{1}{c|}{58.3} & N/A  & \multicolumn{1}{c|}{33.8} & \multicolumn{1}{c|}{N/A} & N/A  \\ \hline
B7          & \multicolumn{1}{c|}{32.8} & \multicolumn{1}{c|}{58.3} & N/A  & \multicolumn{1}{c|}{36.5} & \multicolumn{1}{c|}{N/A} & N/A  \\ \hline\hline
B8          & \multicolumn{1}{c|}{31.8} & \multicolumn{1}{c|}{57.3} & 57.3  & \multicolumn{1}{c|}{37.9} & \multicolumn{1}{c|}{50.2} & 62.2  \\ \hline
B9          & \multicolumn{1}{c|}{32.2} & \multicolumn{1}{c|}{56.8} & 57.6  & \multicolumn{1}{c|}{38.8} & \multicolumn{1}{c|}{48.7} & 62.7  \\ \hline\hline
\end{tabular}
\end{adjustbox}
\caption{Results of the baseline models. B1, B2, and B3 represent the NMT models trained on the original ID data, $Top5$, and $Top6$, respectively. B4, ..., B7 represent models fine-tuned in previous studies. B8 and B9 represent the models trained on the mixture of (ID, $Top5$) and (ID, $Top6$), respectively.}
\label{tbl:baselines}
\end{table}

\subsection{Results and Analysis}
\label{sec:results}
The performance of our fine-tuned MT systems is evaluated with respect to two test sets using case insensitive BLEU~\cite{Papineni2002}, TER~\cite{snover-etal-2006-study} and chrF2~\cite{popovic-2015-chrf} metrics, as implemented withing the sacreBLEU toolkit~\cite{post-2018-call}. Our results are reported in Table~\ref{tbl:result}). We also analyzed the results of different combinations with respect to statistical differences~(see Section~\ref{sec:statistically_significant}).

According to Table~\ref{tbl:result}, the best-to-worst ranking of fine-tuning combinations for both $Top5$ and $Top6$ according to BLEU, TER and chrF2 are $C3, C1, C9, C11, C2, C10, C4, C8, C6, C7, C5$. In our experiments, $C3$ achieved the highest BLEU score among all combinations; except in two cases where other combinations achieved the same scores as follows: (1) $C1$ on test set 2010 for translation of $Top5$ and $Top6$; and (2) $C9$ on test set 2010 and 2011 for translation of $Top5$. 

$C3$, $C1$, and $C9$ also achieved the highest chrF2 and lowest TER scores, suggesting the best setting for fine-tuning our in-domain model could be a combination of (i) a BPE model created from the initial models' training data (i.e., the one used for training models – $top5$ or $Top6$); and (ii) a vocabulary set created from the fine-tuning data (i.e., $ID$); or the initial's models training data or the combination thereof. It is noteworthy that fine-tuning a model using this setting is not always feasible as we may not have access to the data used to train the original model. But, in case of availability, it is suggested to follow the $C3$'s setting.

The next promising fine-tuning setting is $C11$ which suggests combining the initial models' training data with the fine-tuning data. However, if they both are available, we prefer to follow the $C3, C1$, or $C9$ as these may have a better impact on the translation quality. 

The fifth, sixth, and seventh suggested fine-tuning settings according to the evaluation metrics are $C2$, $C10$, and $C4$, respectively. These combinations indicated a model fine-tuned with (i) a BPE model created from the fine-tuning data (i.e., $ID$), and (ii) a vocabulary set created from either the fine-tuning data (i.e., $ID$) or the initial's models training data or the combination thereof could be employed to have an effective fine-tuning process. However, fine-tuning a model with $C10$ and $C4$ is only feasible if one has access to the data used to train models. That is, if the only available data is the one used to fine-tune a model, then it is recommended to follow $C2$.

The other suggested fine-tuning settings according to their evaluation scores in descending order are: $C8, C6, C7, C5$. It is worth mentioning that these setups used both the fine-tuning data (i.e., $ID$) and the initial models' training data, however, did not perform well in terms of translation quality. That shows the importance of BPE models and vocabulary for fine-tuning.

After analyzing our results and extracting a comprehensive guideline of data, BPE, and vocabulary for fine-tuning models, we compared combinations with three categories of baselines to show the effectiveness of the fine-tuning itself regardless of the different fine-tuning combinations. According to the results summarised in Table~\ref{tbl:baselines} and Table~\ref{tbl:result}, all fine-tuned models outperformed the baselines. For example, C3~\footnote{Top5$\leftarrow$ID and evaluated with test set 2010} BLEU scores were increased by roughly 13$\%$ (31.9 to 36.1), 16.8$\%$ (30.9 to 36.1), 10$\%$ (32.8 to 36.1), and 13.5$\%$ (31.8 to 36.1) compared to baselines $B1, B2, B7$, and $B8$ respectively.~\footnote{We chose B7 as the representative of the second baseline category because it achieved the highest BLEU scores among the fine-tuned baselines.} These figures indicate that the fine-tuning process was a better option than training a model at once. These results raise an interesting question about how data should be fed to the neural network at training to achieve optimal performance (both in terms of translation quality as well as training time).

\begin{table*}[t]
\renewcommand{\arraystretch}{1.6}
\centering
\begin{adjustbox}{width=\linewidth,center}
\begin{tabular}{|c|l|l|l|cccccccccccc|}
\hline
\hline
\multirow{4}{*}{\textbf{\#}} &
  \multirow{4}{*}{\textbf{X}} &
  \multirow{4}{*}{\textbf{Y}} &
  \multirow{4}{*}{\textbf{Z}} &
  \multicolumn{12}{c|}{\textbf{Models}} \\ \cline{5-16} 
 &
   &
   &
   &
  \multicolumn{6}{c|}{\textbf{Top5 $\leftarrow$ ID}} &
  \multicolumn{6}{c|}{\textbf{Top6 $\leftarrow$ ID}} \\ \cline{5-16} 
 &
   &
   &
   &
  \multicolumn{3}{c|}{\textbf{Test set 2010}} &
  \multicolumn{3}{c|}{\textbf{Test set 2011}} &
  \multicolumn{3}{c|}{\textbf{Test set 2010}} &
  \multicolumn{3}{c|}{\textbf{Test set 2011}} \\ \cline{5-16} 
 &
   &
   &
   &
  \multicolumn{1}{c|}{\textbf{BLEU$\uparrow$}} &
  \multicolumn{1}{c|}{\textbf{TER$\downarrow$}} &
  \multicolumn{1}{c|}{\textbf{chrF2$\uparrow$}} &
  \multicolumn{1}{c|}{\textbf{BLEU$\uparrow$}} &
  \multicolumn{1}{c|}{\textbf{TER$\downarrow$}} &
  \multicolumn{1}{c|}{\textbf{chrF2$\uparrow$}} &
  \multicolumn{1}{c|}{\textbf{BLEU$\uparrow$}} &
  \multicolumn{1}{c|}{\textbf{TER$\downarrow$}} &
  \multicolumn{1}{c|}{\textbf{chrF2$\uparrow$}} &
  \multicolumn{1}{c|}{\textbf{BLEU$\uparrow$}} &
  \multicolumn{1}{c|}{\textbf{TER$\downarrow$}} &
  \textbf{chrF2} \\ \hline
C1 &
  Top5/6 &
  Top5/6 &
  Top5/6 &
  \multicolumn{1}{c|}{36.1} &
  \multicolumn{1}{c|}{52.4} &
  \multicolumn{1}{c|}{60.3} &
  \multicolumn{1}{c|}{43.7} &
  \multicolumn{1}{c|}{44.1} &
  \multicolumn{1}{c|}{66.1} &
  \multicolumn{1}{c|}{36.4} &
  \multicolumn{1}{c|}{52.3} &
  \multicolumn{1}{c|}{60.4} &
  \multicolumn{1}{c|}{44.1} &
  \multicolumn{1}{c|}{43.7} &
  66.4 \\ \hline
C2 &
  ID &
  ID &
  ID &
  \multicolumn{1}{c|}{35.9} &
  \multicolumn{1}{c|}{52.5} &
  \multicolumn{1}{c|}{60} &
  \multicolumn{1}{c|}{42.8} &
  \multicolumn{1}{c|}{44.9} &
  \multicolumn{1}{c|}{65.2} &
  \multicolumn{1}{c|}{36.1} &
  \multicolumn{1}{c|}{52.5} &
  \multicolumn{1}{c|}{60.0} &
  \multicolumn{1}{c|}{44.0} &
  \multicolumn{1}{c|}{43.7} &
  65.8 \\ \hline
C3 &
  Top5/6 &
  ID &
  Top5/6 &
  \multicolumn{1}{c|}{36.1} &
  \multicolumn{1}{c|}{52.4} &
  \multicolumn{1}{c|}{60.4} &
  \multicolumn{1}{c|}{44.0} &
  \multicolumn{1}{c|}{43.6} &
  \multicolumn{1}{c|}{66.6} &
  \multicolumn{1}{c|}{36.4} &
  \multicolumn{1}{c|}{52.2} &
  \multicolumn{1}{c|}{60.4} &
  \multicolumn{1}{c|}{44.4} &
  \multicolumn{1}{c|}{43.5} &
  66.4 \\ \hline
C4 &
  ID &
  Top5/6 &
  ID &
  \multicolumn{1}{c|}{35.5} &
  \multicolumn{1}{c|}{52.8} &
  \multicolumn{1}{c|}{59.7} &
  \multicolumn{1}{c|}{43.3} &
  \multicolumn{1}{c|}{44.5} &
  \multicolumn{1}{c|}{65.6} &
  \multicolumn{1}{c|}{35.5} &
  \multicolumn{1}{c|}{53.0} &
  \multicolumn{1}{c|}{59.6} &
  \multicolumn{1}{c|}{43.6} &
  \multicolumn{1}{c|}{44.1} &
  65.7 \\ \hline
C5 &
  Top5/6 &
  Top5/6 &
  ID &
  \multicolumn{1}{c|}{33.4} &
  \multicolumn{1}{c|}{53.8} &
  \multicolumn{1}{c|}{58.6} &
  \multicolumn{1}{c|}{40.7} &
  \multicolumn{1}{c|}{45.4} &
  \multicolumn{1}{c|}{64.2} &
  \multicolumn{1}{c|}{33.6} &
  \multicolumn{1}{c|}{53.3} &
  \multicolumn{1}{c|}{58.7} &
  \multicolumn{1}{c|}{40.9} &
  \multicolumn{1}{c|}{45.5} &
  64.2 \\ \hline
C6 &
  ID &
  Top5/6 &
  Top5/6 &
  \multicolumn{1}{c|}{35.4} &
  \multicolumn{1}{c|}{53.0} &
  \multicolumn{1}{c|}{59.7} &
  \multicolumn{1}{c|}{43.5} &
  \multicolumn{1}{c|}{44.6} &
  \multicolumn{1}{c|}{65.5} &
  \multicolumn{1}{c|}{35.5} &
  \multicolumn{1}{c|}{53.6} &
  \multicolumn{1}{c|}{59.9} &
  \multicolumn{1}{c|}{43.3} &
  \multicolumn{1}{c|}{44.4} &
  65.7 \\ \hline
C7 &
  Top5/6 &
  ID &
  ID &
  \multicolumn{1}{c|}{33.4} &
  \multicolumn{1}{c|}{53.3} &
  \multicolumn{1}{c|}{58.6} &
  \multicolumn{1}{c|}{40.9} &
  \multicolumn{1}{c|}{45.1} &
  \multicolumn{1}{c|}{64.2} &
  \multicolumn{1}{c|}{33.2} &
  \multicolumn{1}{c|}{53.4} &
  \multicolumn{1}{c|}{58.2} &
  \multicolumn{1}{c|}{40.2} &
  \multicolumn{1}{c|}{45.4} &
  64.0 \\ \hline
C8 &
  ID &
  ID &
  Top5/6 &
  \multicolumn{1}{c|}{35.4} &
  \multicolumn{1}{c|}{52.8} &
  \multicolumn{1}{c|}{59.6} &
  \multicolumn{1}{c|}{42.8} &
  \multicolumn{1}{c|}{44.6} &
  \multicolumn{1}{c|}{65.1} &
  \multicolumn{1}{c|}{35.1} &
  \multicolumn{1}{c|}{53.1} &
  \multicolumn{1}{c|}{59.6} &
  \multicolumn{1}{c|}{43.3} &
  \multicolumn{1}{c|}{44.4} &
  65.7 \\ \hline
C9 &
    Top5/6 &
    Top5/6+ID &
    Top5/6 &
    \multicolumn{1}{c|}{36.1} &
    \multicolumn{1}{c|}{52.4} &
    \multicolumn{1}{c|}{60.1} &
    \multicolumn{1}{c|}{44.0} &
    \multicolumn{1}{c|}{43.6} &
    \multicolumn{1}{c|}{66.4} &
    \multicolumn{1}{c|}{35.9} &
    \multicolumn{1}{c|}{52.6} &
    \multicolumn{1}{c|}{60.4} &
    \multicolumn{1}{c|}{44.3} &
    \multicolumn{1}{c|}{43.5} &
    66.6 \\ \hline
C10 &
    ID &
    Top5/6+ID &
    ID &
    \multicolumn{1}{c|}{35.6} &
    \multicolumn{1}{c|}{52.7} &
    \multicolumn{1}{c|}{59.9} &
    \multicolumn{1}{c|}{43.0} &
    \multicolumn{1}{c|}{44.9} &
    \multicolumn{1}{c|}{65.2} &
    \multicolumn{1}{c|}{36.1} &
    \multicolumn{1}{c|}{52.3} &
    \multicolumn{1}{c|}{60.2} &
    \multicolumn{1}{c|}{44.0} &
    \multicolumn{1}{c|}{43.6} &
    66.0 \\\hline
C11 &
    Top5/6+ID &
    Top5/6+ID &
    Top5/6+ID &
    \multicolumn{1}{c|}{36.0} &
    \multicolumn{1}{c|}{52.5} &
    \multicolumn{1}{c|}{60.1} &
    \multicolumn{1}{c|}{44.1} &
    \multicolumn{1}{c|}{43.6} &
    \multicolumn{1}{c|}{65.9} &
    \multicolumn{1}{c|}{36.4} &
    \multicolumn{1}{c|}{52.3} &
    \multicolumn{1}{c|}{60.3} &
    \multicolumn{1}{c|}{44.0} &
    \multicolumn{1}{c|}{43.8} &
    66.4 \\\hline \hline
\end{tabular}
\end{adjustbox}
\caption{Evaluation scores of the fine-tuned NMT systems. X represents the source of the BPE model used to create vocabulary; Y represents the source of the vocabulary set and Z represents the source of the BPE model used to create the data for fine-tuning. With $TopN \leftarrow ID$ we denote that a model trained on $TopN$ is fine-tuned on $ID$ data.}
\label{tbl:result}
\end{table*}

\section{Discussions}
\label{sec:discussion}
In this section, first we discuss the pairwise statistical significance of the evaluation scores between the fine-tuned NMT models. Second, we investigate the training time with fine-tuning compared to the training time of baselines. Third, we show to what extent the choice of an initial model for fine-tuning affects the performance of translation.

\subsection{Statistical Significance Test}
\label{sec:statistically_significant}
We computed pairwise statistical significance of the results shown in Table~\ref{tbl:result} in terms of BLEU scores by using bootstrap resampling and 95\% confidence interval for both test sets (2010 and 2011) based on 1000 iterations, and samples of 300 sentences. According to the test output, most fine-tuned models have a statistically significant difference except those systems pairs listed in Table \ref{tbl:ss_exception}. 

In addition to the analysis presented in Section~\ref{sec:results}, this shows two main points: (1) if there is no access to both initial and fine-tuned models' data we can achieve quite similar performance only from the initial data. For example, the initial model –$Top5$– fine-tuned as per $C1$ and $C3$ on the 2010 test set have no differences in terms of BLEU score. (2) having one single data set versus two different ones for creating BPE models and vocabulary may not always have a significant impact on the the model performance. For example, while $C2$ was trained with one single data set (ID), $C4$ employed both ID and Top5 and still achieved the same performance (on test set 2010).

\begin{table}[ht]
\renewcommand{\arraystretch}{1.2}
\centering
{\small \setlength\tabcolsep{1.8pt} 
\begin{tabular}{l|c|c|}
\cline{2-3} \cline{2-3} 
                                     & \textbf{Test Set 2010} & \textbf{Test Set 2011} \\ \hline
\multicolumn{1}{|r|}{\textbf{Top5}}  & \begin{tabular}[c]{@{}c@{}}(C1, C3) \\ (C6, C8)\\ (C2, C4) \\ (C5, C7) \end{tabular}      & \begin{tabular}[c]{@{}c@{}}(C1, C6) \\ (C4, C5) \end{tabular}      \\ \hline
\multicolumn{1}{|r|}{\textbf{Top6}} & \begin{tabular}[c]{@{}c@{}}(C1, C3) \\ (C6, C8) \end{tabular} & \begin{tabular}[c]{@{}c@{}} (C1, C6) \\ (C4, C5) \end{tabular} \\ \hline \hline
\end{tabular}}
\caption{Results of systems pairs that are not statistically significant (for $p<0.05$). (CX, CY) means models that fined tuned with the setup suggested in combinations CX and CY have no statistically significant difference based on 300 samples.}
\label{tbl:ss_exception}
\end{table}

\subsection{Training Time}
In Table~\ref{tab:training_time} we present the running time (RT) for training the baselines ($B1, B2, B3, B8$ and $B9$) and for fine-tuning for the best model ($C3$) for both $Top5$ and $Top6$.\footnote{Fine-tuning for all models and SW and vocabulary combinations took approximately 1h and 30 minutes. As such we limit our discussion to $C3$.} Fine-tuning time is about 1 hour and 30 minutes compared to the training time of an MT system with all data at once ($B8$ or $B9$) which is about 5 or 6 hours. That shows that the time for fine-tuning is only a fraction (27\%) of the time for training models on all data at once. On the one hand, compared to the sum of training and fine-tuning times, that is B2 or B3 followed by C3 which amount at 12 hours and 6 minutes and 9 hours and 52 minutes accordingly, training a model ``from scratch'' is preferable. On the other hand, training ``from scratch'' does not achieve the same performance as with fine-tuning (as already stated this in Section~\ref{sec:results}).

\begin{table}[ht]
\renewcommand{\arraystretch}{1.3}
\centering
\begin{adjustbox}{width=\linewidth,center}
\begin{tabular}{|l|c|l|c|l|}
\hline
\hline
\multicolumn{1}{|l|}{\textbf{\#}} &
\textbf{\begin{tabular}[c]{@{}c@{}}Complete RT\\ D:H:M\end{tabular}} &
\multicolumn{1}{c|}{\textbf{Step}} &
\multicolumn{1}{l|}{\textbf{\begin{tabular}[c]{@{}c@{}}Best model RT\\ D:H:M\end{tabular}}}  &
\multicolumn{1}{l|}{\textbf{Step}} \\ 
\hline
B1   & 00:03:53   & 18,000 & 00:00:50   & 5,000   \\
B2   & 00:10:33   & 35,000 & 00:05:50   & 20,000  \\
B3   & 00:08:20   & 35,000 & 00:04:26   & 19,000  \\ 
\hline \hline
B8   & 00:05:47   & 23,000 & 00:03:16   & 13,000 \\
B9   & 00:06:17   & 25,000 & 00:04:45   & 15,000 \\
\hline \hline
C3-Top5 &  00:01:33   & 12,000 & 00:00:17   & 2,000   \\
C3-Top6 &  00:01:32   & 12,000 & 00:00:16   & 2,000   \\
\hline \hline

\end{tabular}
\end{adjustbox}
\caption{Running time (RT) for training and fine-tuning. The first baselines B1, B2 and B3 are trained on original in-domain data, $Top5$, and Top6 accordingly; baselines B8 and B9 are combinations of ID and Top5 or ID and Top6. $C3-Top5$ and $C3-Top6$ indicate the fine-tuning with the best combination ($C3$) of models trained on $Top5$ and $Top6$ accordingly.}
\label{tab:training_time}
\end{table}

\subsection{Reverse Fine-tuning }
We also assess whether the initial model for fine-tuning in our case study ($Top5/6)$ is effective or we may need to reverse the order in which data is presented for training. That is training MT systems on the original in-domain data followed by fine-tuning them on the selected in-domain data. It is noteworthy that we conducted this experiment to monitor the performance as well as the sensitivity of the fine-tuning process toward the initial model. 

According to Table~\ref{tbl:reverse_FT}, all fine-tuned models outperformed the reverse fine-tuned models. That means, starting with the selected in-domain models ($Top5/6$) is more efficient than starting with the original in-domain. This also reveals the fact that trained on large data (such as $Top5/6$) prevents overfitting on the small in-domain data (such as $ID$). This transformation to the new parameters can cause a drop in models performance for the test instances from the initial data~\cite{b2ea90abe88b4eb983ba5315000e28db,dakw:17fine}. However, this is not entirely true in our case study as we worked on one domain for the entire research. That is, both initial and fine-tuning data are from one specific domain, and only the first one is relatively large. In the future we plan to expand on more domains and assess the impact of different data quantities and domain-specificity.

\begin{table}[h]
\renewcommand{\arraystretch}{1.5}
\centering
\begin{adjustbox}{width=\linewidth,center}
\begin{tabular}{|c|ccc|ccc|}
\hline \hline
\multirow{2}{*}{\#} & \multicolumn{3}{c|}{Test set 2010}                            & \multicolumn{3}{c|}{Test set 2011}                            \\ \cline{2-7} 
                   & \multicolumn{1}{c|}{BLEU$\uparrow$} & \multicolumn{1}{c|}{TER$\downarrow$}  & CHRF2$\uparrow$ & \multicolumn{1}{c|}{BLEU$\uparrow$} & \multicolumn{1}{c|}{TER$\downarrow$}  & CHRF2$\uparrow$ \\ \hline
$ID\leftarrow Top5$ & \multicolumn{1}{c|}{30.8} & \multicolumn{1}{c|}{59.4} & 56.8 & \multicolumn{1}{c|}{36.3} & \multicolumn{1}{c|}{52.3} & 61.1 \\ \hline
$ID\leftarrow Top6$          & \multicolumn{1}{c|}{31.5} & \multicolumn{1}{c|}{58.2} & 57.3  & \multicolumn{1}{c|}{37.8} & \multicolumn{1}{c|}{50.3} & 61.9  \\ \hline\hline
\end{tabular}
\end{adjustbox}
\caption{The results of reverse fine-tuning. With $ID \leftarrow TopN$ we denote that a model trained on $I$ is fine-tuned on $TopN$ data.}
\label{tbl:reverse_FT}
\end{table}

\section{Related Work}
\label{sec:related_work}
There is a significant amount of research on the topic of fine-tuning MT. Most prior studies have investigated adapting models to a different domain. That is, they first employed a large out-of-domain data and then fine-tuned it on small in-domain data. 
Among others, \cite{luong-etal-2015-effective} did the first successful work, where they trained a model on English-German general-domain data and then fine-tuned it on a new domain data (conversational) in the same languages. They claimed an increase of 3.8 BLEU points compared to the original model (25.6 to 29.4) without further training.

Another method to improve the translation performance on the new domain without degrading the performance on the generic domain test set was proposed in \cite{freitag2016fast}. To this end, they ensembled the fine-tuned model with the already trained baseline model; and evaluated their method by IWSLT 2015 evaluation campaign~\cite{cettolo-etal-2015-iwslt}. The authors reported a gain of 7.2 BLEU points on the in-domain test set while still retaining the performance on the out-of-domain test set. 

Following that, \cite{freitag2016fast} demonstrated that while an ensemble approach for fine-tuning seems a good option, the performance of the fine-tuned models still drops for the generic domain task, especially when it comes to domain-specific contexts (e.g., medical and legal domain). This is mainly because the in-domain data set at topic or genre level~\cite{van-der-wees-etal-2015-whats}, comprises new vocabulary and linguistic features that are different from the generic data~\cite{koehn-knowles-2017-six}. To fix this problem, they proposed a fine-tuning method based on knowledge distillation~\cite{https://doi.org/10.48550/arxiv.1503.02531}.

There is also other research carried out to solve the vocabulary mismatches in the context of fine-tuning. For instance, \cite{sato-etal-2020-vocabulary} proposed a method to adapt the embedding layers of the initial model to the target domain. They performed this by projecting the general word embedding obtained from target-domain monolingual data onto source-domain embedding. The authors reported 3.86 and 3.28 BLEU points gain in English$\rightarrow$Japanese and German$\rightarrow$English translation, respectively.

As segmenting words and generating vocabulary hugely impact the MT performance \cite{ataman-federico-2018-evaluation} extensively evaluated the problem of segmenting words at a subword level and compared two word segmentation methods: Byte-Pair Encoding (BPE) ~\cite{sennrich-etal-2016-neural} and the Linguistically-Motivated Vocabulary Reduction (LMVR)~\cite{Ataman2017} for NMT. They compared these approaches in five morphologically-rich languages and reported that LMVR achieved better performance in the tested languages.

\section{Conclusion and Future Work}
\label{sec:conclusion}
In this paper, we conducted a systematic analysis based on a commonly used domain-specific data set (IWSLT 2014) to find the optimal combination of subword generation approach and vocabulary for fine-tuning NMT models. In addition to comparing the performance of 22 models (11 options for two training sets), we investigated how fine-tuning impacts training time and MT performance, compared to training an MT model with all data at once. Through our empirical evaluation, we have created a state-of-the-art model for in-domain translation that can be employed in different contexts, among others, multilingual domain adaptation~\cite{cooper-stickland-etal-2021-multilingual}.

Considering the available data, we present two decision points (i.e., subwords and vocabulary) that need to be made prior to fine-tuning, along with a third one about crossing the BPE source for the training/fine-tuning data and for the vocabulary. 
Our experiments outline a roadmap with three possible options for fine-tuning in-domain models as follows: (1) In case both the initial model's data~($D$) and fine-tuning data~($E$) are available, it might be effective to train BPE models and create vocabulary using $D$ and $E$, respectively. (ii) Otherwise, it might be viable to create BPE and vocabulary based on $D$; and (iii) if the last two options were not possible, it suggests creating the decision points all based on $E$. 
It is worth noting that all fine-tuning strategies the initial MT models improved the baselines, with the maximum gained of 6 BLEU points (30.9 to 36.1). 

In our future work, we intend to improve the generalization of our pre-trained (in-domain) models by further training on an out-of-domain corpus, so it possibly enables these models to translate generic inputs as well as their specialized context without forgetting what they already learned. Another research direction would be to investigate the proposed decision points on other language pairs and domains.

The data and fine-tuned models are available at:~\url{https://github.com/JoyeBright/FT-IWSLT2014-BPEVocab}.

\bibliographystyle{\conffilename}
\bibliography{\conffilename}

\end{document}